\documentclass[sigconf]{aamas}

\usepackage{caption}
\usepackage{subcaption}

\usepackage{balance} % for balancing columns on the final page
\usepackage{quoting}
\usepackage{mdframed}
\usepackage{orcidlink}

\setcopyright{ifaamas}
\acmConference[AAMAS '24]{Proc.\@ of the 23rd International Conference
on Autonomous Agents and Multiagent Systems (AAMAS 2024)}{May 6 -- 10, 2024}
{Auckland, New Zealand}{N.~Alechina, V.~Dignum, M.~Dastani, J.S.~Sichman (eds.)}
\copyrightyear{2024}
\acmYear{2024}
\acmDOI{}
\acmPrice{}
\acmISBN{}

\acmSubmissionID{1250}

\title{Conversational Language Models for Human-in-the-Loop Multi-Robot Coordination}
\subtitle{Demonstration Track}

\author{William Hunt\, 
\orcidlink{0000-0003-4269-5050}}
\affiliation{
  \institution{University of Southampton}
  \city{Southampton}
  \country{United Kingdom}}
\email{W.Hunt@soton.ac.uk}

\author{Toby Godfrey\,
\orcidlink{0009-0004-4501-5051}}
\affiliation{
  \institution{University of Southampton}
  \city{Southampton}
  \country{United Kingdom}}
\email{tmag1g21@soton.ac.uk}

\author{Mohammad D. Soorati\, 
\orcidlink{0000-0001-6954-1284}}
\affiliation{
  \institution{University of Southampton}
  \city{Southampton}
  \country{United Kingdom}}
\email{M.Soorati@soton.ac.uk}

\begin{abstract}
With the increasing prevalence and diversity of robots interacting in the real world, there is need for flexible, on-the-fly planning and cooperation. Large Language Models are starting to be explored in a multimodal setup for communication, coordination, and planning in robotics. Existing approaches generally use a single agent building a plan, or have multiple homogeneous agents coordinating for a simple task. We present a decentralised, dialogical approach in which a team of agents with different abilities plans solutions through peer-to-peer and human-robot discussion. We suggest that argument-style dialogues are an effective way to facilitate adaptive use of each agent's abilities within a cooperative team. Two robots discuss how to solve a cleaning problem set by a human, define roles, and agree on paths they each take. Each step can be interrupted by a human advisor and agents check their plans with the human. Agents then execute this plan in the real world, collecting rubbish from people in each room. Our implementation uses text at every step, maintaining transparency and effective human-multi-robot interaction.
\end{abstract}


\newcommand{\BibTeX}{\rm B\kern-.05em{\sc i\kern-.025em b}\kern-.08em\TeX}

\makeatletter
\gdef\@copyrightpermission{
	\begin{minipage}{0.3\columnwidth}
		\href{https://creativecommons.org/licenses/by/4.0/}{\includegraphics[width=0.90\textwidth]{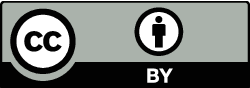}}
	\end{minipage}\hfill
	\begin{minipage}{0.7\columnwidth}
		\href{https://creativecommons.org/licenses/by/4.0/}{This work is licensed under a Creative Commons Attribution International 4.0 License.}
	\end{minipage}
	\vspace{5pt}
}
\makeatother

\begin{document}
%%% The following commands remove the headers in your paper. For final 
%%% papers, these will be inserted during the pagination process.

\pagestyle{fancy}
\fancyhead{}

%%% The next command prints the information defined in the preamble.

\maketitle

\renewenvironment{quote}[1][\unskip]
 {\vspace{0.2cm}
  \begin{mdframed}[backgroundcolor=#1] 
  \small\list{}{\rightmargin=-0.5cm \leftmargin=-0.45cm}%
  \item\relax
  \quoting
 }
 {\endquoting
 \end{mdframed}
 \endlist
 \vspace{0.2cm}
}

%%%%%%%%%%%%%%%%%%%%%%%%%%%%%%%%%%%%%%%%%%%%%%%%%%%%%%%%%%%%%%%%%%%%%%%%

\section{Introduction}
Multirobot systems, while still reasonably rare in everyday life, are becoming increasingly common in domains such as agriculture~\cite{lytridis2021overview, albiero2022swarm}, search and rescue~\cite{clark2022industry}, and construction~\cite{werfel2011distributed}. It is projected that the household robotics market will grow considerably this decade~\cite{householdRep}, leading to large numbers of robots interacting not only with their users but also with each other. This presents increasing demand for robotic systems which can flexibly adapt to new tasks without prior training. There is also a recent trend towards generalist robotics; the same model or overarching structure can be deployed on a wide variety of platforms and tasks~\cite{open_x_embodiment_rt_x_2023}. To this end, it is desirable to move towards a domain-agnostic coordination structure that is common across platforms so that the group can be treated as a single entity. A popular area of recent development in the AI landscape is that of Large Language Models (LLMs)~\cite{yang2023harnessing}; agents that operate in the domain of text through next-symbol prediction. LLMs may present an effective way to interpret task or agent descriptions, as well as calling on internalised understanding and reasoning to develop the available information. LLMs can also facilitate some degree of communication between agents that helps them organise while keeping the control flow understandable and accessible to a human operator. Multi-modal AI development is now influencing robotics and other agents-based research fields~\cite{driess2023palm} by using LLMs to power end-to-end approaches which can understand and use human-written inputs  to inform their actions~\cite{brohan2023can}. This fits into an overarching vision towards a multimodal, generalist agent which can understand and operate on text, images, and other inputs in robotics~\cite{reed2022generalist}. Although conversational agents such as ChatGPT are typically used to model a human-agent conversation, some works have focused on modelling a conversation between multiple agents. This is typically done through ``role-playing''; an agent is told to ``imagine'' that it is a person with a certain role and then enters a conversation whilst assuming that role~\cite{li2023camel}. This process can be used to model internal monologue for a single agent who ``talks to themself'' about what they can perceive and do~\cite{huang2022inner}. Conversational approaches can also use multiple personas, each of whom brings a different specialised perspective to the collective generation of text by editing the group solution to fulfill different goals that they each have for the end result~\cite{wang2023unleashing}.

Role-playing has been used for a variety of tasks including debate~\cite{chan2023chateval}, auctions, haggling~\cite{nascimento2023self, fu2023improving}, and checking with a human supervisor that they are not hallucinating~\cite{ren2023robots}. These approaches set the conditions for a dialogue and leave the agents to talk, assuming that an intelligent solution emerges naturally from the conversation. This has been used to create a team of software developers who each write code and pass responsibilities to the next developer~\cite{hong2023metagpt}. LLMs have also been integrated into robotic simulators for communication to organise who performs each task, allowing robots to coordinate their workspaces and decide which of them is able to reach an object~\cite{mandi2023roco}. A similar approach simulates agents fetching items in a house, they communicate and incorporate short-term memory to request assistance from each other on the fly~\cite {zhang2023building}. Some works include diverse skillsets where each agent has different capabilities and skills, such as a work where the agents build a team with the required skills before planning and acting~\cite{chen2023agentverse}.

We present a proof-of-concept system that leverages the knowledge of pretrained models by building language-based agents which talk to each other, and with humans, using natural language. This allows agents to discuss and debate their strategies towards a collaborative solution to a high-level mission objective with observation or assistance from a human supervisor. This forms a pipeline that allows agents to take a high-level task description and autonomously perform every step of the process, from planning to assignment.

\section{Demonstration}
\begin{figure}
     \centering
     \begin{subfigure}[b]{0.22\textwidth}
         \centering
         \includegraphics[height=2.1cm]{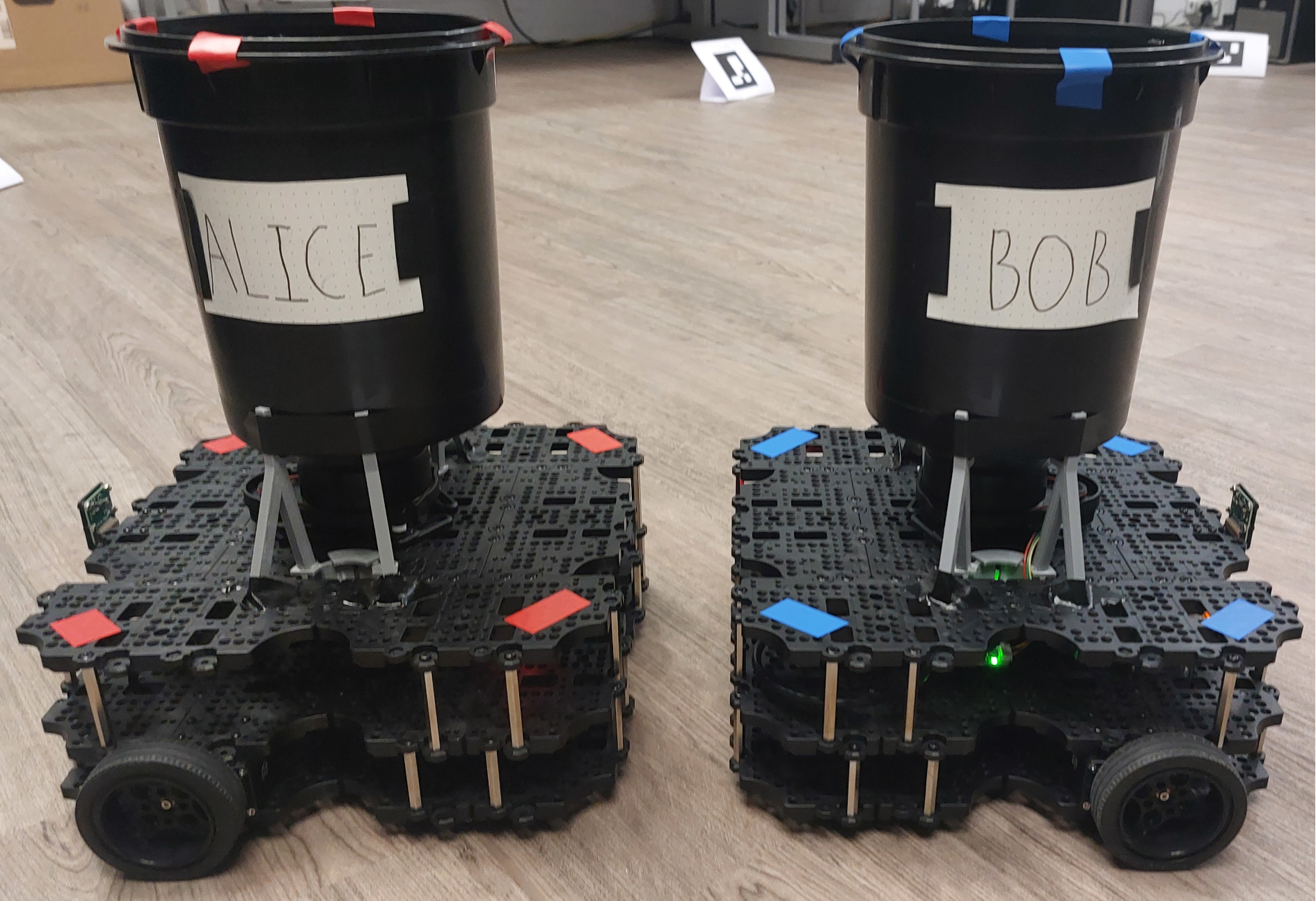}
         \vspace{-1mm}
         \caption{Cleaning Robots}
         \label{fig:robots}
     \end{subfigure}
     \hfill
     \begin{subfigure}[b]{0.22\textwidth}
         \centering
         \includegraphics[height=2.1cm]{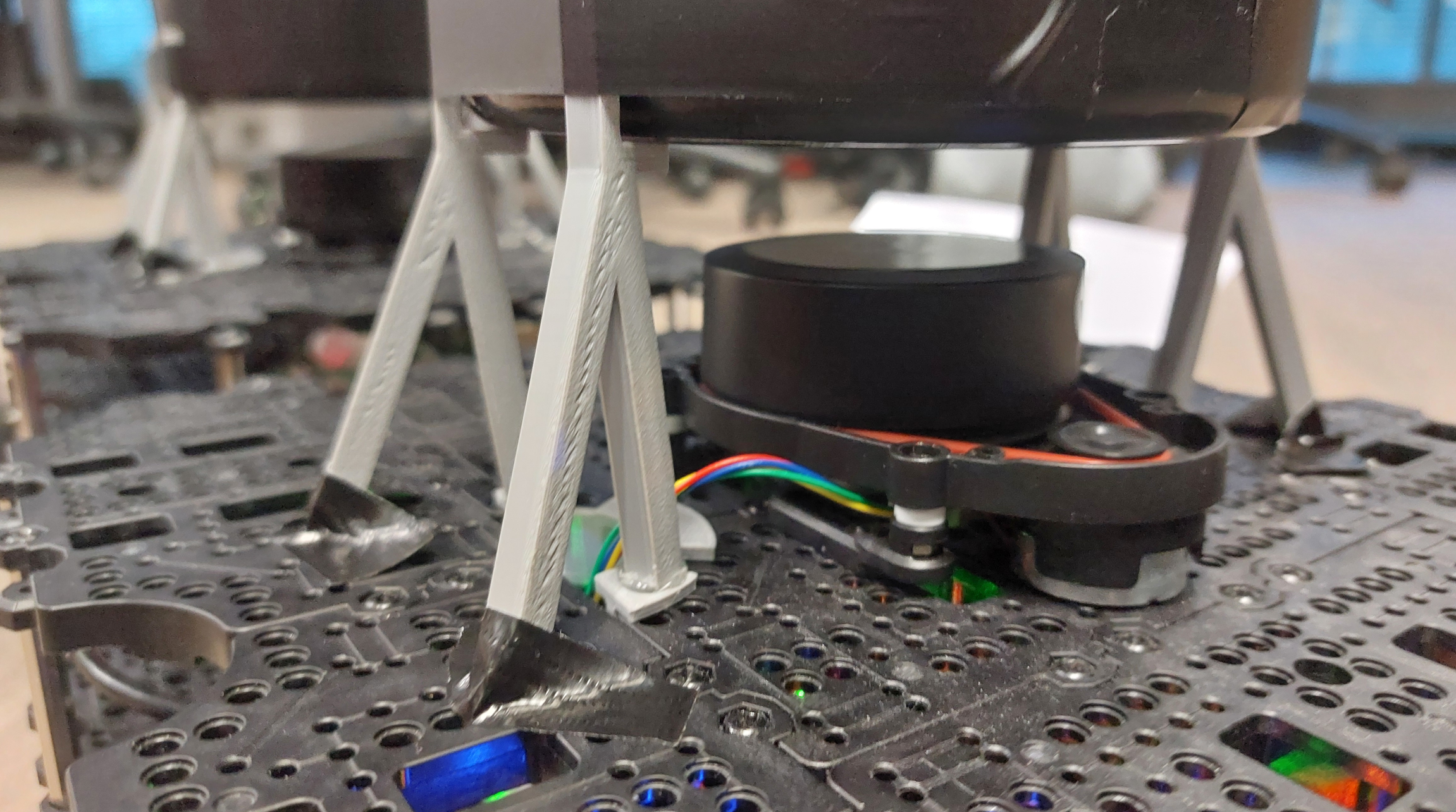}
         \vspace{-1mm}
         \caption{3D printed connection}
         \label{fig:connection}
     \end{subfigure}

     \begin{subfigure}[b]{0.22\textwidth}
         \centering
         \includegraphics[height=2.1cm]{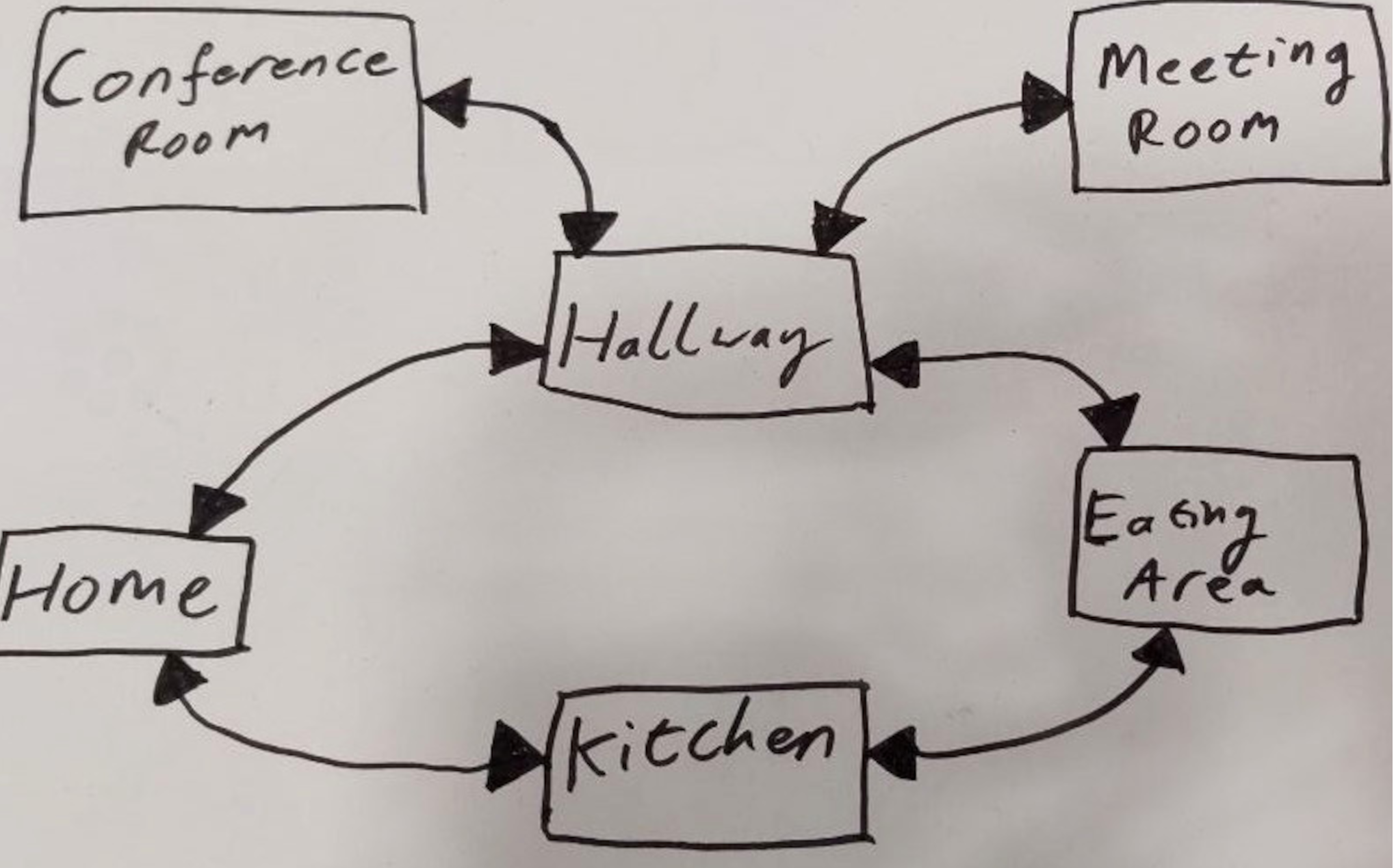}
         \vspace{-1mm}
         \caption{Map of environment}
         \label{fig:diagram}
     \end{subfigure}
    \hfill
     \begin{subfigure}[b]{0.22\textwidth}
         \centering
         \includegraphics[height=2.1cm]{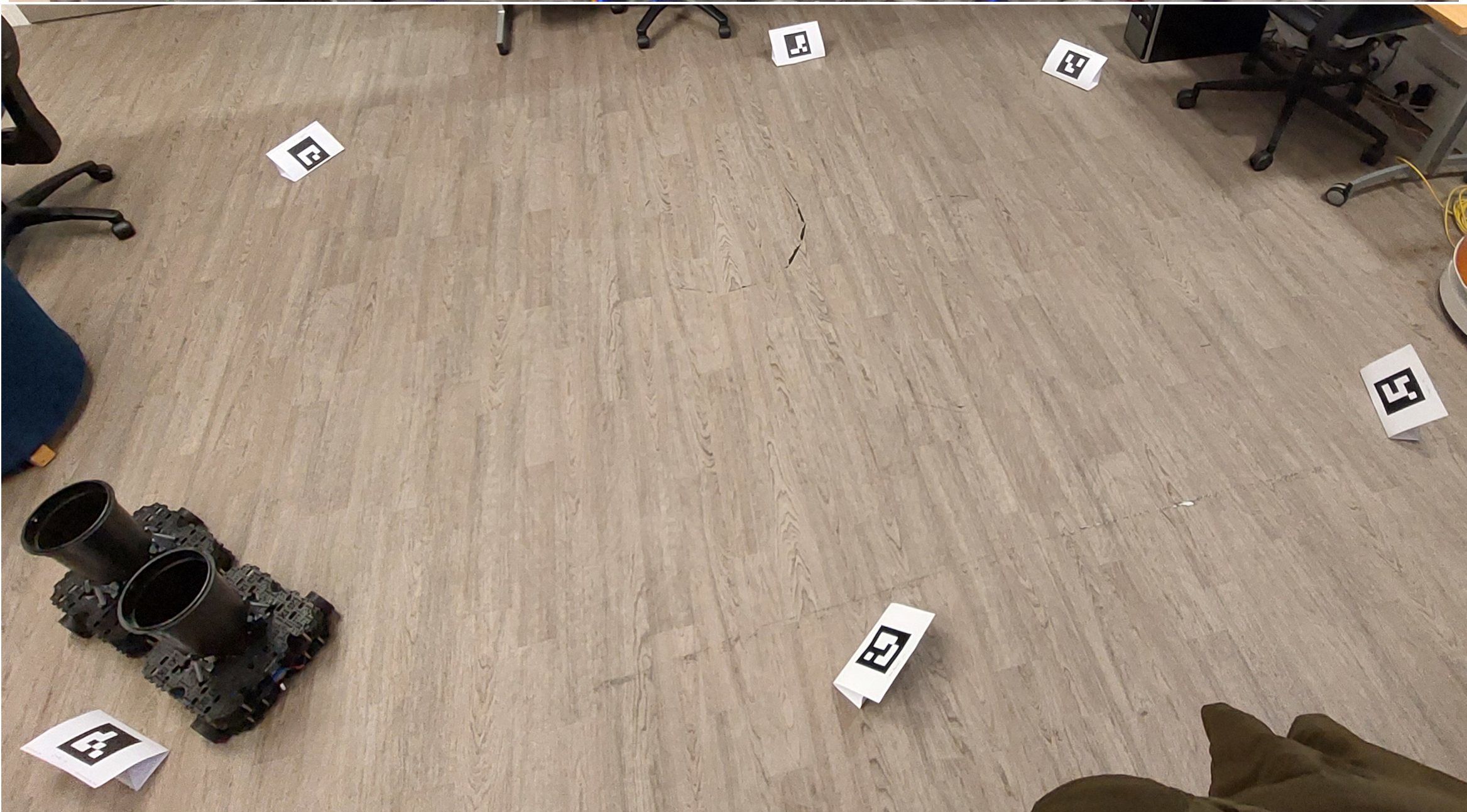}
         \vspace{-1mm}
         \caption{View of arena}
         \label{fig:arena}
     \end{subfigure}
        \vspace{-2mm}
        \caption{Cleaning Robots and the demonstration setup.}
        \label{fig:modelDiagram}
\end{figure}

An LLM is used to create a conversation where agents build paths to be executed on hardware, which in turn can detect problems and prompt the LLM to re-plan. The pipeline uses text at every step of the process to retain deep meaning from end to end. A Python program takes human input, and calls the GPT API (``gpt-4-vision-preview'')~\cite{OpenAI_API} for language generation. The conversation produces plans which are passed directly to the robots. In brief, the steps are: \textbf{(1) Agent Ego}: The ``system message'' (identity) for each agent is set to a description of each agent, plus some general guidance on how to debate; \textbf{(2) Environment Description}: We provide a flowchart-style environment model that shows agents which rooms are connected with arrows (see Fig.\ref{fig:diagram}); \textbf{(3) Human Supervisor}: The supervisor presents a task to the agents; \textbf{(4) Discussion}: Agents discuss the task and plan their approach; \textbf{(5) Calling the supervisor}: Agents call the supervisor when done, or if they need help. The supervisor can ask for alterations if desired or approve the plan.
%In brief, the steps are:
%\begin{enumerate}
%    \item \textbf{Agent Ego}: The ``system message'' (identity) for each agent is set to a description of each agent, plus some general guidance on how to debate.
%    \item \textbf{Environment Description}: We provide a flowchart-style environment model that shows agents which rooms are connected with arrows.
%    \item \textbf{Human Supervisor}: The supervisor presents a task to the agents.
%    \item \textbf{Discussion}: Agents discuss the task and plan their approach.
%    \item \textbf{Calling the supervisor}: Agents call the supervisor when done, or if they need help. The supervisor can ask for alterations if desired or approve the plan.
%\end{enumerate}
Agents are added to the chat and a human supervisor is given a chat box to speak with them. When agents discuss, they start each message with their name and it is added to the log. From each agent's perspective, they perceive every other agent's message as one from the human, but the name tags allow them to understand the conversation properly. The system prompt, which the LLM considers with every message, encourages agents to negotiate with each other. This is important because the default GPT configuration is polite and rarely contradicts its interlocutor, however for collective planning agents should point out mistakes. When they reach a decision, a path of rooms for each agent is extracted and passed to hardware for execution. Two TurtleBot3 robots (see Fig.\ref{fig:robots}) are controlled with ROS2 Humble on a Raspberry Pi 4, they are each equipped with a LIDAR and an optical camera. ROS gathers data from the LIDAR and controls the differential-drive motors. The LIDAR is used for collision avoidance to protect the hardware as it moves around the environment. The optical camera is used to detect ArUco markers which indicate rooms.

\begin{figure}
    \centering
    \includegraphics[width=\linewidth]{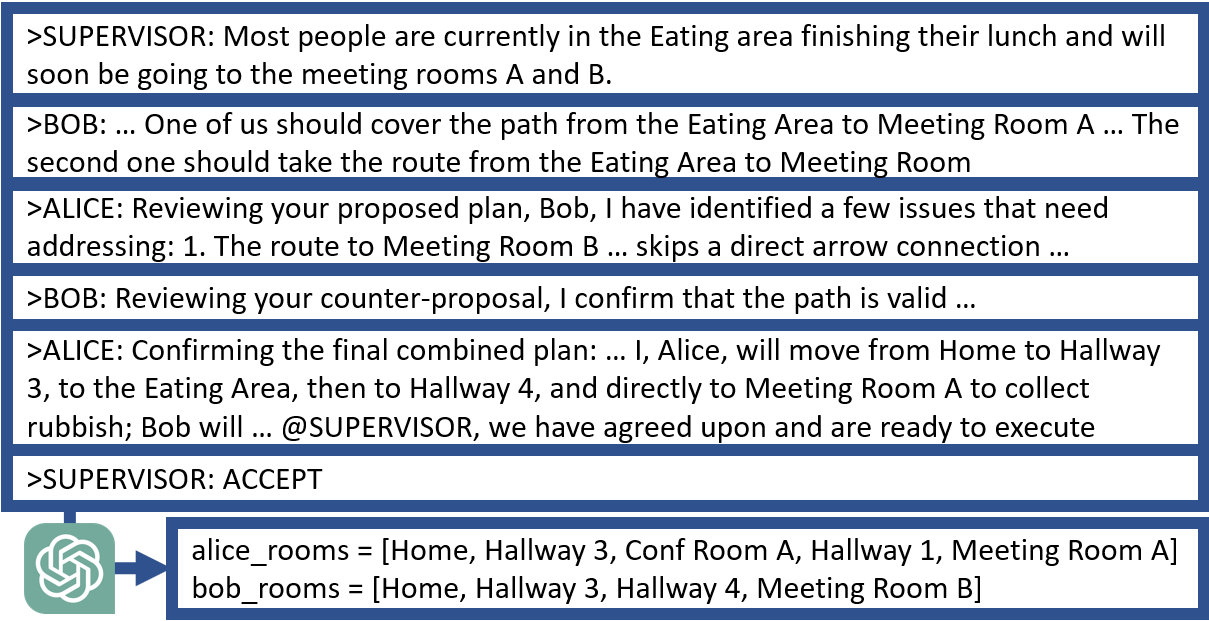}
    \vspace{-5mm}
    \caption{An example dialogue where the rooms are extracted.}
    \label{fig:systemDiagram}
\end{figure}

%\section{Demonstration}
We demonstrate a conversational planning system deployed on real robots to simulate autonomous waste collection. A bin is mounted on top of each robot using a 3D-printed structure (see Fig.\ref{fig:connection}), turning the robots into mobile bins (see Fig.\ref{fig:robots}). Participants can engage by typing a task into the supervisor PC, and then watching the agents converse. The participants can also offer advice to the robots or point out issues before approving the plan. The robots then move around the arena to solve the task in real time. 
Participants are encouraged to provide vague and unusual instructions or interrupt the execution with novel information to test the system and explore the challenges of the approach as well as its potential. A cut-down example of this process is shown in Fig.\ref{fig:systemDiagram}, where the agents must clean up after lunch at a conference. They define two roles, correct a mistake, and decide on the division of labour.
\footnote{Our demo video can be found here: \url{https://www.youtube.com/watch?v=cVCwG8aLIvI}}

\section{Conclusion}
Robotic systems would benefit greatly from being able to understand, interpret, and utilise otherwise ignored text data. Inter-agent conversation may be a useful tool for decentralised mission planning with a human in the loop. We demonstrate conversational multi-agent coordination that allows agents to be represented with few-word names and calls on the deep knowledge of Large Language Models. The language-based approach can leverage expert opinion across many scenarios as the entire system is understandable and can be interfaced directly by a human user as much as is required. The proposed demonstration uses a language model to allow two robots to plan and execute a garbage collection task with a human supervisor in the loop and a large screen will display the conversations. Participants can interact and interrupt the system with different textual inputs to learn more about the capabilities and limitations of using language models for multi-robot coordination. 

%%%%%%%%%%%%%%%%%%%%%%%%%%%%%%%%%%%%%%%%%%%%%%%%%%%%%%%%%%%%%%%%%%%%%%%%

%%% The acknowledgments section is defined using the "acks" environment
%%% (rather than an unnumbered section). The use of this environment 
%%% ensures the proper identification of the section in the article 
%%% metadata as well as the consistent spelling of the heading.

\begin{acks}
This project was done as part of the Fast-PI project funded by the UKRI Trustworthy Autonomous Systems Hub [EP/V00784X/1]. It was also supported by UK Research and Innovation [EP/S024298/1].
\end{acks}

%%%%%%%%%%%%%%%%%%%%%%%%%%%%%%%%%%%%%%%%%%%%%%%%%%%%%%%%%%%%%%%%%%%%%%%%
\balance
%%% The next two lines define, first, the bibliography style to be 

\bibliographystyle{ACM-Reference-Format} 
\bibliography{sample}

%%%%%%%%%%%%%%%%%%%%%%%%%%%%%%%%%%%%%%%%%%%%%%%%%%%%%%%%%%%%%%%%%%%%%%%%

\end{document}